\pdfoutput=1

\documentclass[11pt]{article}

\usepackage{acl}

\usepackage{times}
\usepackage{latexsym}
\usepackage{multirow}
\usepackage{subcaption}
\usepackage{microtype}
\usepackage{booktabs}
\usepackage{url}
\usepackage{eurosym}
\usepackage{todonotes}
\usepackage{amsmath}
\usepackage{enumitem}
\usepackage{array}
\usepackage{longtable}
\usepackage{makecell}
\usepackage{tabularx}
\usepackage{algorithm} 
\usepackage{algpseudocode}
\usepackage{soul,xcolor}
\usepackage{framed}
\usepackage{cleveref}
\usepackage{graphicx}

\newcommand{\figref}[1]{Figure~\ref{fig:#1}}

\newcommand{\tabref}[1]{Table~\ref{tab:#1}}
\newcommand{\finqa}{\text{FinQA}}
\newcommand{\tatqa}{\text{TAT-QA}}
\newcommand{\model}{\textsc{TAT-LLM}}

\newcommand\fone{F\textsubscript{1}}

\usepackage[T1]{fontenc}

\usepackage[utf8]{inputenc}

\usepackage{microtype}

\title{TAT-LLM: A Specialized Language Model for Discrete Reasoning over Tabular and Textual Data}

\author{Fengbin Zhu$^{1,3}$, Ziyang Liu$^{3}$, Fuli Feng$^{2}$, Chao Wang$^{3}$, Moxin Li$^{1}$, Tat-Seng Chua$^1$\\
\textsuperscript{1}National University of Singapore, Singapore\\~\textsuperscript{2}University of Science and Technology of China, China\\~\textsuperscript{3}6Estates Pte Ltd, Singapore\\
}

\begin{document}
\maketitle
\begin{abstract}

In this work, we address question answering (QA) over a hybrid of tabular and textual data, involving a variety of common content in reality like SEC filings, where discrete reasoning is often required.
We consider harnessing the multi-step reasoning capabilities of large language models (LLMs) to tackle this problem, which have recently achieved remarkable success in many natural language tasks.
To do this, we first abstract a \emph{Step-wise Pipeline} for tabular and textual QA to help LLMs better execute multi-step inference, containing three key steps of \textit{Extractor}, \textit{Reasoner} and \textit{Executor}.
We initially design an instruction to validate the pipeline on GPT-4, demonstrating promising results. 
However, utilizing an online LLM like GPT-4 holds various challenges in terms of cost, latency, and data security risk, which motivates us to specialize smaller LLMs in this task.
We then develop a \model~model by fine-tuning LLaMA 2 with the training data generated automatically from existing datasets following the \emph{Step-wise Pipeline}.
The experimental results have verified that our \model~model can outperform all compared models, including prior best fine-tuned models and very large-scale LLMs like GPT-4 on FinQA, TAT-QA and TAT-DQA benchmarks.
The models and datasets are publicly available at \url{https://huggingface.co/next-tat} 
\end{abstract}

\section{Introduction}

\begin{figure}[t]
    \begin{center}
    \includegraphics[scale=0.68]{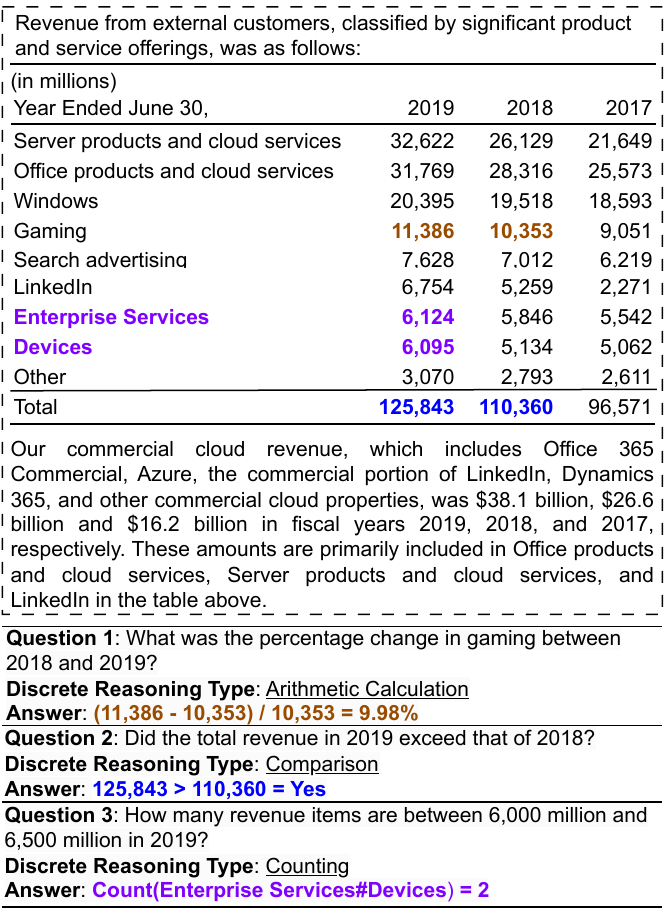}
    \end{center}
    \vspace{-0.4cm}
    \caption{Examples of QA with discrete reasoning over a hybrid of tabular and textual data.}
    \label{fig:sample}
    \vspace{-1.6em}
\end{figure}

The documents containing both tables and text, e.g. SEC filings, academic papers and medical reports, make a very prevalent category of content in the real world. 
They often feature extensive numerical data in both the tabular and textual content, necessitating discrete reasoning capabilities for machines to comprehend them.
Recent research~\cite{zhu2021tat,chen2021finqa} investigates the intelligent comprehension of such documents through question answering (QA) tasks, as exemplified in Figure~\ref{fig:sample}. 
The model, provided with a table and relevant text as the context, needs to perform various types of discrete reasoning, such as arithmetic calculations, making comparisons, and counting, to answer the question.

To perform QA over hybrid tabular and textual data, a straightforward approach~\cite{ran2019numnet} involves taking the table, text, and question as input and generating the answer directly.
This approach can be ineffective due to the complex reasoning process involved~\cite{wei2022chain}. 
To address this issue, some works~\cite{lei2022answering,zhou2022unirpg,Zhu2023SoarGraph} decompose the task into multiple steps, producing intermediate results that serve as references for the final answer. 
These multi-step approaches typically design distinct modules at each step and often optimize these modules concurrently through multi-task learning. 
To date, there has been no consensus on how to decompose the answer process in existing literature.

\begin{figure*}[t]
    \begin{center}
    \includegraphics[scale=0.8]{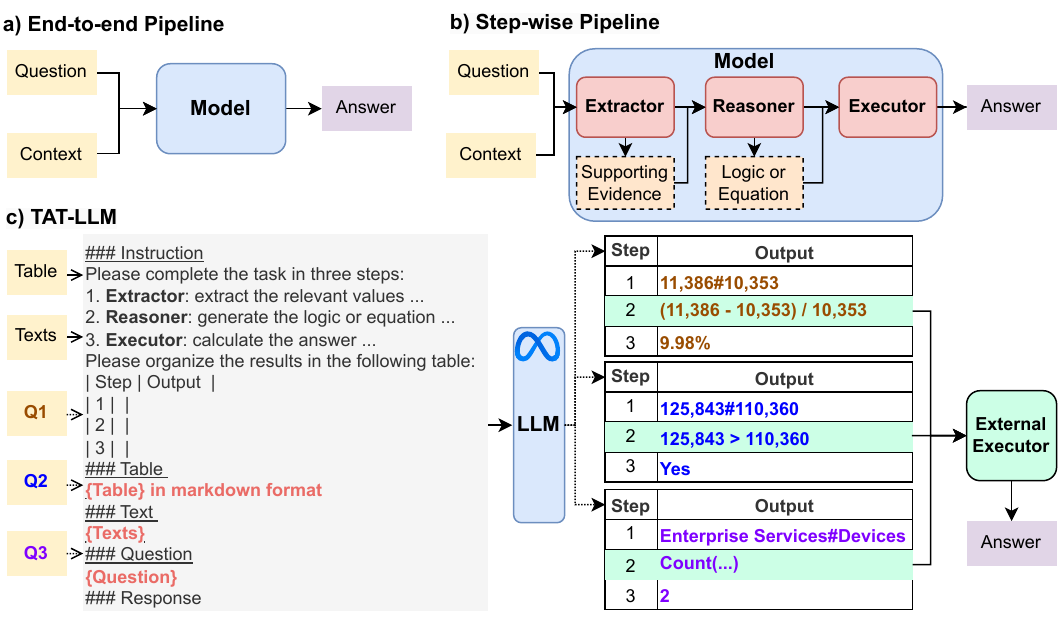}
    \end{center}
    \caption{\label{fig:qa_system} 
Comparison between a) \emph{End-to-end Pipeline} and b) \emph{Step-wise Pipeline}. c) Our \model~ language model is developed by fine-tuning LLaMA 2 following the \emph{Step-wise Pipeline}. }
  \vspace{-1em}
\end{figure*}

Recently, large language models (LLMs) like GPT-4~\cite{openai2023gpt4} and FLAN~\cite{wei2022finetuned} have exhibited strong multi-step reasoning abilities~\cite{wei2022emergent} with proper instructions such as chain-of-thought (CoT)~\cite{wei2022chain} and least-to-most~\cite{zhou2023leasttomost}. 
Therefore, we consider harnessing this amazing power of LLMs for better discrete reasoning over hybrid tabular and textual data.
To achieve this, we first identify three key steps in the process of tabular and textual QA from previous multi-step  methods~\cite{zhu2021tat,Zhu2023SoarGraph,zhou2022unirpg}, and abstract a \emph{Step-wise Pipeline}, as illustrated in \figref{qa_system} b).
Specifically, 1) \textit{Extractor} identifies the relevant information or evidence to the question from the given context; 2) \textit{Reasoner} generates a mathematical equation or logic rule  with the obtained information; and 3) \textit{Executor} derives the final answer by executing the mathematical equation or logic rule with the associated information.
The three steps emphasize different capabilities of the tabular and textual QA model --- understanding the question and context, inferring the logic for answering the question, and calculating the answer with precision.
These steps produce a sequence of intermediate results, which means we can specifically model and enhance one (or more) of them given a specific application scenario.

Following the \emph{Step-wise Pipeline}, we initially design a task instruction and validate it on GPT-4~\cite{openai2023gpt4}, achieving promising results on multiple benchmarks. 
However, utilizing an online LLM presents challenges in terms of cost, latency, and data security risk. 
By contrast, fine-tuning a smaller language model, specifically for math word problems~\cite{fu2023specializing,cobbe2021gsm8k}, has been proven fairly appealing.
We are then motivated to explore the specialization of smaller language models for addressing this challenge following the \emph{Step-wise Pipeline}.

We develop a \model~model by fine-tuning LLaMA 2~\cite{touvron2023llama2} with the training data generated automatically from existing expert-annotated datasets, as shown in \figref{qa_system} c).
In particular, we train the selected LLM to take the table, text, and question as input, and complete the task in three key steps following the \emph{Step-wise Pipeline}. 
We construct each of the training instances with five parts: Instruction, Table, Text, Question, and Response, from available tabular and textual QA datasets.
Inspired by Tab-CoT~\cite{ziqilu2023tabcot}, the model output (or response) is formatted to a structured table, with each row corresponding to one step of the pipeline, as shown in the right part of \figref{qa_system} c).
This design is friendly to further refining the outputs at each step and automatic evaluation.
Moreover, to enhance the model performance, we equip the \model~model with an \emph{External Executor}, which strengthens the execution of logical rules and mathematical calculations to better infer the final answer.
We test our \model~on three popular benchmarks: FinQA~\cite{chen2021finqa}, TAT-QA~\cite{zhu2021tat} and TAT-DQA~\cite{zhu2022towards}.
The experimental results show that our smallest model \model~(7B) can outperform all baseline models and even beat GPT-4 on all three datasets. 
In summary, we make the following main contributions in this work:
\begin{itemize}[leftmargin=*]
    \item We abstract a \emph{Step-wise Pipeline} including \textit{Extractor}, \textit{Reasoner} and \textit{Executor} to assist LLMs in better performing discrete reasoning over a hybrid of tabular and textual data.
    \item We develop a \model~model by fine-tuning LLaMA 2 (including $7$B, $13$B and $70$B) for tabular and textual QA following the \emph{Step-wise Pipeline}, better supporting practical application with cost and privacy concerns. 
    \item We conduct extensive experiments on three popular benchmarks, validating the superiority of our \model~model over both conventional methods and very large-scale LMs, e.g. GPT-4.
\end{itemize}

\section{Related Work}

\subsection{Tabular and Textual QA}

Early works on tabular and textual QA~\cite{chen2020hybridqa,li2021tsqa,chen2020open} focus on answer extraction from tabular and textual data. 
In recent two years, a growing body of works tackle tabular and textual QA by performing discrete reasoning as massive numerical information is usually included in the table or text, such as TAT-QA~\cite{zhu2021tat}, TAT-HQA~\cite{li2022learning}, FinQA~\cite{chen2021finqa}, MultiHiertt~\cite{zhao2022multihiertt}, and TAT-DQA~\cite{zhu2022towards}.
In this line of research, many supervised fine-tuning methods are proposed, such as DyRRen~\cite{Li2023DyRRen}, RegHNT~\cite{lei2022answering}, UniRPG~\cite{zhou2022unirpg} and MVGE~\cite{wei2023multiview}.
Harnessing the capabilities of advanced LLMs to tackle tabular and textual QA remains a relatively underexplored area, which is the main focus of our work.

\subsection{Large Language Models (LLMs)}
Recently, LLMs like ChatGPT \cite{ouyang2023following} have attracted tremendous research attention with their superb language understanding and generation abilities, bringing excellent performance to many tasks like reading comprehension~\cite{10.1145/3560260}, discrete reasoning~\cite{wei2022chain}, etc. 
Since these general LLMs are usually extremely large and not accessible for most researchers, open-source smaller LLMs are proposed to allow researchers to customize their own LLMs for different tasks, such as LLaMA \cite{touvron2023llama} and Alpaca \cite{alpaca}, with comparable performance to the general LLMs. 
Recently, there has been a trend in the research community that small LLMs are specialized for specific tasks through fine-tuning with instructions~\cite{fu2023specializing}, yielding impressive performance.
To the best of our knowledge, we are the first work to specialize a small LLM (e.g.,7B) in tabular and textual QA challenge.

\section{Approach}
In this section, we first introduce the \emph{Step-wise Pipeline} for tabular and textual QA, and then elaborate our proposed \model~model.

\subsection{\emph{Step-wise Pipeline}}

Large language models (LLMs) like ChatGPT~\cite{ouyang2023following} and FLAN~\cite{wei2022finetuned} have demonstrated astoundingly strong multi-step reasoning abilities~\cite{wei2022emergent} following human instructions in many tasks~\cite{zhou2023leasttomost,cobbe2021gsm8k}.
To better leverage such amazing power of LLMs to address QA over hybrid tabular and textual data, we abstract a \emph{Step-wise Pipeline} from the previous multi-step methods~\cite{zhu2021tat,zhou2022unirpg} on this task. 
As shown in \figref{qa_system} b), this pipeline addresses the task with three key steps: \textit{Extractor}, \textit{Reasoner}, and \textit{Executor}.
In particular, \textit{Extractor} serves as an information extraction module that identifies the relevant snippets or segments of information to the question from the context;  
2) \textit{Reasoner} works as a ``logic thinker'' to generate the right logic rule or equation that leads to the right answer to the question;
3) \textit{Executor} is responsible for deriving the final answer by performing the logic rule or executing the equation.
Compared with an \emph{End-to-end Pipeline} that provides little insight into how the answer is derived, as shown in \figref{qa_system} a),
such a \emph{Step-wise Pipeline} is able to produce a sequence of intermediate results, which means we can boost each component (or some of them) within the pipeline to consequently acquire improved final performance.
For example, in this work, we choose to strengthen the third step by adding an extra Executor, which brings noticeable performance gains as verified experimentally in Section \ref{executor_exp}.

To instruct LLMs to perform discrete reasoning over tabular and textual data following the \emph{Step-wise Pipeline}, we carefully design a natural language instruction to include the three key steps in the \emph{Step-wise Pipeline}.
Please refer to Appendix~\ref{appe:template_zero} for more information.
We validate it on both FinQA and TAT-QA benchmarks using ChatGPT~\cite{ouyang2023following} and GPT-4~\cite{openai2023gpt4} and obtain promising results, demonstrating the rationality of applying the \emph{Step-wise Pipeline} grounded on the LLMs.
We then generate training data with publicly available tabular and textual QA datasets and specialize a smaller LLM, i.e., LLaMA 2~\cite{touvron2023llama2}, to answer questions following the \emph{Step-wise Pipeline} for better supporting practical applications with cost and privacy concerns, as elaborated at below.

\subsection{TAT-LLM}

\paragraph{Selection of Language Model.}
We develop our \model~ by fine-tuning  LLaMA 2 ~\cite{touvron2023llama2}.
LLaMA \cite{touvron2023llama} is a series of open-source large language models trained on the large-scale publicly available corpus.
It is commonly employed in the development of large language models with instruction tuning, such as Alpaca~\cite{alpaca} and Vicuna~\cite{vicuna2023}.
In this study, we choose the latest LLaMA 2~\cite{touvron2023llama2} as our base language model considering it offers an extended maximum sequence length (i.e., $4,096$) compared to the previous version LLaMA~\cite{touvron2023llama}, which often fails to accommodate the tabular and textual content of one instance from existing datasets.
Moreover, LLaMA 2 has demonstrated remarkable performance across various benchmarks.

\begin{algorithm}[t]
\caption{: External Executor}
\textbf{Input} $O_1$: the output of \textit{Extractor}; 
$O_2$: the output of \textit{Reasoner}; $O_3$: the output of \textit{Executor}; $Q_t$: the predicted question type. 
    \begin{algorithmic}[1]
    \State $answer \gets O_3$
    \If{$O_2$ is a valid arithmetic equation} 
        \State $answer\gets round(eval(O_2), 4)$
    \ElsIf{``\#'' in $O_2$} \# multiple values are separated with ``\#''
            \State $arr \gets O_2.split(``\#")$ 
            \State $answer \gets len(arr)$
    \ElsIf{``>'' in $O_2$ or ``<'' in $O_2$} 
           \State $answer \gets eval(O_2)$
    \ElsIf{$O_2$  = ``N.A.''}
        \If{$Q_t$ = ``Span''}
            \State $answer \gets O_1$
        \ElsIf{$Q_t$ = ``Multiple Spans''}
            \State $arr \gets O_1.split(``\#")$ \# spans are separated with ``\#''
            \State $answer \gets arr$
        \EndIf
    \EndIf
    \end{algorithmic}
\label{algo}
\end{algorithm}

\paragraph{Construction of Training Data.}
Our fine-tuning is based on the training data we automatically transform from existing tabular and textual QA datasets.
For implementation, we use three datasets, i.e., FinQA~\cite{chen2021finqa}, TAT-QA~\cite{zhu2021tat} and TAT-DQA~\cite{zhu2022towards}.
We carefully design templates for each dataset to generate the training data for fine-tuning LLMs. 
Please refer to Appendix~\ref{appe:template_finetuning} for more information.
As shown in \figref{qa_system} c), each training instance is designed to be composed of five parts,  including Instruction, Table, Text, Question, and Response.

\noindent $\bullet$ \textbf{Instruction:}
This part provides a detailed guide for what the task is and how to complete it. 
It outlines the steps that should be followed to derive the answer from the provided context. 
We adopt the \emph{Step-wise Pipeline} described in above sections.
1) \textit{Extractor}: We request the model to extract the relevant information from either the table or the accompanying text. 
2) \textit{Reasoner}: Based on the extracted information, we ask the model to derive an equation or a logic rule that is used to infer the answer.
3) \textit{Executor}: We require the model to execute the equation or logic rule to arrive at the final answer.
An illustration is given in  \figref{qa_system} c), which is the same for all the training instances. 
The instruction part helps the model to precisely understand the task and generate accurate and meaningful outputs.

\noindent $\bullet$ \textbf{Table}: This part refers to the tabular data in the input context of this task.
We compose this part using the tables from FinQA and TAT-QA datasets. 
As each table in the two datasets is stored in a Two-Dimensional (2D) array, we transform it into a markdown table.
Note that we omit TAT-DQA dataset for this part, considering the context given in TAT-DQA is the document pages from PDF files, where the structure of the table is unknown.

\noindent $\bullet$ \textbf{Text}: This part refers to the textual data in the input context.
For FinQA and TAT-QA datasets, we compose this part with the paragraphs relevant to the table; all contents from the document pages in TAT-DQA dataset are regarded as textual data.

\noindent $\bullet$ \textbf{Question}: This part refers to the question that is asked based on the given tabular and textual context. 
We directly put the question from the three datasets in this part. 
The goal of the task is to generate the right answer to this question.

\begin{table}[t]
    \centering
    \large
    \begin{tabular}{l|ccc}
    \toprule
       \bf Dataset & \bf Train Set& \bf Dev Set  & \bf Test Set \\
       \midrule
        FinQA & 6,251 & 883 & 1,147 \\
        TAT-QA & 13,215 & 1,668 & 1,669 \\
        TAT-DQA & 13,251 & 1,645 & 1,662 \\
     \bottomrule
\end{tabular}
     \caption{Statistics of the train, validation and test splits for the financial tabular and textual QA datasets. }
      \label{tab:dataset}
 \vspace{-1.6em}
\end{table}
\noindent $\bullet$ \textbf{Response}: This part describes the model's output based on the given table, text, and question. 
To better inspect the results in the intermediate steps and automatically evaluate the model output, we train the language model to format the output in a structured markdown table with two columns, i.e., step and output.
Each row in the output table corresponds to one step defined in the instruction.
In the end, the model draws its conclusion with a statement: "The answer is: \{answer\}", making its final response clear.

To train the model, the response part in the training set needs to be generated in advance. 
Existing tabular and textual QA datasets often provide the answer to the question and the derivation annotated by human experts.
To generate the correct response for constructing our training instance, we first automatically identify the supporting evidence from the derivation of each question, such as the numbers and entities, which are used as the output of the first step.
We then convert the derivation to a valid equation or logic rule as the output of the second step,  making sure its execution result is consistent with the final answer.
We directly use the annotated answer as the final answer, which is also the output of the last step.

\paragraph{Training.}
After constructing the training data, we train our \model~model in various sizes, including 7B, 13B, and 70B, by fine-tuning LLaMA 2~\cite{touvron2023llama2} using Low-Rank Adaptation (LoRA)~\cite{hu2022lora}.
More details are provided in Appendix~\ref{Implementation Details}.

\paragraph{External Executor.}
We observe that the trained model struggles in performing the execution of the mathematical equations and logic rules (e.g., counting, comparison) in the last step, according to experimental evaluations in Section \ref{executor_exp}.
Such incompetence of the Executor has also been found in previous works~\cite{luo2023wizardmath, yue2023mammoth}.
To enhance the model's accuracy of the final output, we propose to add an \emph{External Executor} to our model, which takes the model's intermediate outputs as input and performs the execution of the equations or logic rules to obtain the final answer. 
With this \emph{External Executor}, the model can refine its output to a better one.
With the output of the model formatted to a structured table, we can easily access the intermediate results.  
After obtaining the intermediate results, the \emph{External Executor} is applied to refine the final answer instead of directly using the prediction of the model.
The whole process is summarized in Algorithm \ref{algo}.

\section{Experiments}

\begin{table}[t]
    \centering 
\begin{tabular}{llc}
\toprule
\bf Type & \bf Model & \bf EM \\
\midrule
\multicolumn{2}{l}{\bf Human Expert Performance} &  91.16 \\
\midrule
\multirow{7}{*}{\bf \rotatebox[origin=c]{90}{Fine-tuned} } & Longformer & 21.90 \\
& NeRd & 48.57 \\
& $\text{FinQANet}_\textit{BERT}$ & 50.00 \\
& $\text{DyRRen}_\textit{BERT}$ & 59.37 \\
& $\text{FinQANet}_\textit{RoBERTa}$ & 61.24 \\
& $\text{ELASTIC}_\textit{RoBERTa}$ & 62.66 \\
& $\text{DyRRen}_\textit{RoBERTa}$ & 63.30 \\
\midrule
\multirow{7}{*}{\bf \rotatebox[origin=c]{90}{Zero-shot}} & Vicuna (7B) & 10.11 \\
& LLaMA 2-Chat (7B) & 15.43 \\
& LLaMA 2-Chat (70B) & 32.17 \\
& MAmmoTH (70B) & 36.09 \\
& WizardMath (70B) & 47.25 \\
& GPT3.5-Turbo & 58.00 \\
& GPT-4 & \underline{63.91} \\
\midrule
\multirow{2}{*}{ \bf Ours} 
& \multirow{2}{*}{\model~(7B)} & (\textcolor{red}{ +0.69})  \\
&  & \bf 64.60   \\

 \bottomrule
\end{tabular}
     \caption{Performance of our \model~model and compared models on the test set of \finqa. 
     Best results are marked in bold and numbers in red indicate the improvement over the underlined second-best results.
     }
      \label{tab:result_finqa}
 \vspace{-1.6em}
\end{table}
\subsection{Datasets, Models and Evaluation Metrics}
\label{datasets}
\paragraph{Datasets.} 
We use FinQA ~\cite{chen2021finqa}, TAT-QA \cite{zhu2021tat} and TAT-DQA \cite{zhu2022towards} for our experiments.
See \tabref{dataset} for the statistics of splits of each dataset.

\noindent $\bullet$ \textbf{FinQA}~\cite{chen2021finqa} is an expert-annotated tabular and textual QA dataset in which the tables and text are sampled from financial reports. It focuses on discrete reasoning capabilities like addition, subtraction, multiplication, division, and numerical comparison. 

\noindent $\bullet$ \textbf{TAT-QA}~\cite{zhu2021tat} is also built with tables and paragraphs extracted from financial reports.
Most questions require discrete reasoning to generate the answers, and meanwhile, there are also cases where the answers can be extracted directly from the tables or text.
Its questions are classified into four different types: \textit{Span}, \textit{Multiple Spans}, \textit{Counting}, and \textit{Arithmetic}.

\noindent $\bullet$ \textbf{TAT-DQA}~\cite{zhu2022towards} is an extension of the TAT-QA dataset, focusing on question answering over the original long financial statements with up to three pages, and the position and structure of the tables are unknown.

\begin{table}[t]
    \centering
     \small
\begin{tabular}{llcc}
\toprule
\bf Type & \bf Model & \bf EM & \bf \fone \\
\midrule
\multicolumn{2}{l}{\bf Human Expert Performance} &  84.1 &	90.8 \\
\midrule
\multirow{11}{*}{\bf \rotatebox[origin=c]{90}{Fine-tuned} } & TagOp & 50.10 & 58.00 \\
& TeaBReaC & 55.80 & 63.80 \\
& KIQA & 58.20 & 67.40 \\
& FinMath & 58.30 & 68.20 \\
& GANO & 61.90 & 72.10 \\
& MHST & 63.60 & 72.70 \\
& UniPCQA & 63.90 & 72.20 \\
& SoarGraph & 65.40 & 75.30 \\
& UniRPG & 67.20 & 76.00 \\
& RegHNT & 70.30 & 77.90 \\
& MVGE & 70.90 & 79.10 \\
\midrule
\multirow{7}{*}{\bf \rotatebox[origin=c]{90}{Zero-shot}} & Vicuna (7B) & 32.53 & 40.97 \\
& LLaMA 2-Chat (7B) & 37.16 & 45.37 \\	
& MAmmoTH (70B) & 38.97 & 46.51 \\
& WizardMath (70B)  & 39.63 &	45.28 \\
& LLaMA 2-Chat (70B) & 45.94 & 53.80 \\
& GPT3.5-Turbo & 59.47 & 68.11 \\
& GPT-4 & \underline{71.92} & \underline{79.71} \\
\midrule
\multirow{2}{*}{ \bf Ours} 
& \multirow{2}{*}{\model~(7B)} & (\textcolor{red}{+2.64})  & (\textcolor{red}{+3.17}) \\
&  & \bf 74.56 & \bf 82.88 \\

 \bottomrule
\end{tabular}
 \caption{Performance of our \model~model and compared models on the test set of \tatqa. 
     }
      \label{tab:result_tatqa}
  \vspace{-2em}
\end{table}

\paragraph{Compared Models.} 
We compare our \model~model with two kinds of models: fine-tuned models on the tabular and textual QA dataset, and LLMs in zero-shot setting. 
For fine-tuned models, we select the state-of-the-art supervised fine-tuned models for each dataset, which are listed separately in \tabref{result_finqa}, \tabref{result_tatqa} and \tabref{result_tatdqa}.
For LLMs, we utilize the state-of-the-art GPT3.5-Turbo~\cite{openai2020gpt3} and GPT-4 \cite{openai2023gpt4}.
Besides, we also adopt powerful smaller LLMs that have garnered considerable research interest in the field, including Vicuna (7B) \cite{vicuna2023} and LLaMA 2-Chat (7B and 70B) \cite{touvron2023llama2}. 
In addition to these general LLMs, we also compare our models with LLMs specialized in math word problems, including MAmmoTH (70B)~\cite{yue2023mammoth} and WizardMath (70B) \cite{luo2023wizardmath}, since our task involves numerical reasoning. 
All LLMs are tested in zero-shot setting,
because financial tabular and textual data inputs tend to be lengthy, making it impractical to include additional in-context examples due to input length limits.

\paragraph{Evaluation Metrics.} For TAT-QA and TAT-DQA datasets, we adopt the Exact Match (EM) and the numeracy-focused (macro-averaged) \fone{} score \cite{zhu2021tat, zhu2022towards}. 
Both two metrics measure the overlap between a bag-of-words representation of the gold and predicted answers. 
The numeracy-focused \fone{} score is set to 0 unless the predicted number is exactly equal to the ground truth.
We use EM for FinQA, which is the same as the metric of Execution Accuracy originally used in FinQA~\cite{chen2021finqa}.

For zero-shot prediction using LLMs, we omit the scale prediction and compare the predicted value with the ground truth value only, assuming the scale prediction is always correct. 
For our \model~model, since we instruct the model to output a structured table that is friendly to automatic evaluation, we take the scale into account.

\begin{table}[t]
    \centering
     \small
\begin{tabular}{llcc}
\toprule
\small
\bf Type & \bf Model & \bf EM & \bf \fone \\
\midrule
\multicolumn{2}{l}{\bf Human Expert Performance} &  84.1 &	90.8 \\
\midrule
\multirow{4}{*}{\bf \rotatebox[origin=c]{90}{Fine-tuned} } & 
NumNet+ V2 &  30.60 & 40.10\\
& TagOp & 33.70 & 42.50\\
& MHST & 41.50 & 50.70 \\
& Doc2SoarGraph & 59.20 & 67.60 \\
\midrule
\multirow{7}{*}{ \bf \rotatebox[origin=c]{90}{Zero-shot} } & Vicuna (7B) & 28.44 & 36.72 \\
& LLaMA 2-Chat (7B) & 34.52 & 42.32 \\
& MAmmoTH (70B) & 35.42 &	42.82 \\
& WizardMath (70B) & 36.44 &	41.55 \\
& LLaMA 2-Chat (70B) & 41.91 & 49.74 \\	
& GPT3.5-Turbo & 52.74 & 61.40 \\
& GPT-4 & \underline{64.46} & \underline{72.20} \\
\midrule
\multirow{2}{*}{ \bf Ours} 
& \multirow{2}{*}{\model~(7B)} & (\textcolor{red}{ +4.99})  & (\textcolor{red}{+5.55}) \\
&  & \bf 69.45 & \bf 77.75  \\

\bottomrule
\end{tabular}
     \caption{Performance of our \model~ model and compared models on the test set of TAT-DQA. 
     }
    \label{tab:result_tatdqa}
     \vspace{-1.8em}
\end{table}

\subsection{Main Results} 
\label{main_result}
We first compare the performance of our \model~ with previous methods on each dataset respectively. 
The experimental results are summarized in \tabref{result_finqa} for FinQA, \tabref{result_tatqa} for TAT-QA and \tabref{result_tatdqa} for TAT-DQA. 
From the tables, we make the following observations.
1) Our \model~(7B) significantly outperforms all the previous models on each of the three datasets, including the previous best fine-tuned models and the state-of-the-art GPT-4~\cite{openai2023gpt4}. 
In particular, our \model~reaches $64.60\%$,  $74.56\%$ and $69.45\%$ in terms of EM on the test set of FinQA, TAT-QA and TAT-DQA respectively, i.e. an increase of $0.69$, $2.64$ and $4.99$ points compared to GPT-4. These results well demonstrate the noticeable effectiveness of our \model~model.
The results also confirm the rousing potential of specializing smaller language models, like our \model~(7B) model, for specific tasks to yield better performance than very large-scale models like GPT-4.   
2) However, the performance of GPT-4 and our \model~(7B) obviously lags behind that of human experts, showing that this task is still challenging. 
3) The strong general LLM GPT3.5-Turbo~\cite{ouyang2023following} and LLaMA 2-Chat (70B)~\cite{touvron2023llama2} underperform the best fine-tuned models, evidencing that supervised fine-tuning is still essential for achieving advanced performance for this challenge. 
4) The LLMs specialized in mathematical reasoning, i.e., WizardMath~\cite{luo2023wizardmath} and MAmmoTH~\cite{yue2023mammoth}, largely underperform GPT3.5-Turbo, GPT-4 and our \model~model, showing that current numerically-enhanced LLMs still struggle in discrete reasoning over tabular and textual QA.
This again speaks for the considerable value of specializing models for specific tasks.

\begin{table}[t]
    \centering
    \tiny
\begin{tabular}{lccccc}
 \toprule
\multirow{2}{*}{ \bf Model }  & \bf FinQA & \multicolumn{2}{c}{\bf TAT-QA} & \multicolumn{2}{c}{\bf TAT-DQA} \\
 \cmidrule(lr){2-2}
 \cmidrule(lr){3-4}
\cmidrule(lr){5-6}
& \bf EM & \bf EM & \bf \fone &\bf EM & \bf \fone \\
\midrule
GPT-3.5-Turbo & 58.00 & 59.47 & 68.11 & 52.74 & 61.40 \\
GPT-4 & \underline{63.91} & \underline{71.92} & \underline{79.71} & \underline{64.46} & \underline{72.20} \\
\midrule
\multicolumn{6}{l}{\bf{Fine-tuned with respective training set:}}\\
\multirow{2}{*}{\model~\text{(7B)}}  & (\textcolor{red}{+0.69}) &  (\textcolor{red}{+2.64}) & (\textcolor{red}{+3.17}) & (\textcolor{red}{+4.99}) & (\textcolor{red}{+5.55})\\
& 64.60 & 74.56  & 82.88 & 69.45 & 77.75 \\
\midrule
\multicolumn{6}{l}{\bf{Fine-tuned with a combination of training sets:}}\\
\multirow{2}{*}{$\model_\textit{All}$~\text{(7B)}} & (\textcolor{red}{+1.22}) &  (\textcolor{red}{+4.57}) & (\textcolor{red}{+5.42}) & (\textcolor{red}{+6.92}) & (\textcolor{red}{+8.04})\\
 & 65.13  & 76.49 &  85.13 & 71.38 & 80.24  \\
 \cmidrule{2-6}
 
 \multirow{2}{*}{$\model_\textit{All}$~\text{(13B)}} & (\textcolor{red}{+8.02}) &  (\textcolor{red}{+5.59}) & (\textcolor{red}{+6.24}) & (\textcolor{red}{+7.76}) & (\textcolor{red}{+8.36})\\
 & 71.93 & 77.51 & 85.95 & 72.22 & 80.56  \\
 \cmidrule{2-6}

 \multirow{2}{*}{$\model_\textit{All}$~\text{(70B)}}  & (\textcolor{red}{+12.90}) &  (\textcolor{red}{+9.50}) & (\textcolor{red}{+8.78}) & (\textcolor{red}{+12.09}) & (\textcolor{red}{+11.70})\\
 & \bf 76.81 & \bf 81.42 & \bf 88.49 & \bf 76.55 & \bf 83.90\\
 \bottomrule
\end{tabular}
\caption{Performance of the \model~model trained with combination of all three training sets.
     }
    \label{tab:train_all}
      \vspace{-2.5em}
\end{table}

\subsection{In-Depth Analysis}

\paragraph{Fine-tuning with Combined Data.}
In addition to training our \model~model on a single tabular and textual QA dataset, we also train the $\model_{All}$~(7B) with a combination of the train sets from FinQA, TAT-QA and TAT-DQA datasets. 
As shown in \tabref{train_all}, training on a combined dataset brings clear performance improvements over training on a single dataset. 
This suggests that incorporating larger-sized and more diverse data into the training process can potentially enhance the model's overall capabilities. 
Furthermore, we check the performance with various sizes of the base LLaMA~2 model, and find that larger models consistently achieve significant performance improvements across all three datasets, aligning with the prevailing consensus in the field. 
Compared to GPT-4, $\model_{ALL}$ (70B) increases $12.90\%$, $9.50\%$ and $12.09\%$ on the EM over FinQA, TAT-QA and TAT-DQA, respectively, which strongly validates the rationality of our solution.

\begin{table}[t]
    \centering
    \scriptsize
\begin{tabular}{lccccc}
\toprule
\multirow{2}{*}{ \bf Model }  & \bf FinQA & \multicolumn{2}{c}{\bf TAT-QA} & \multicolumn{2}{c}{\bf TAT-DQA} \\
 \cmidrule(lr){2-2}
 \cmidrule(lr){3-4}
\cmidrule(lr){5-6}
& \bf EM & \bf EM & \bf \fone &\bf EM & \bf \fone \\
\midrule
GPT-4 & 63.91 & 71.92 & 79.71 & 64.46 & 72.20 \\
\midrule
\multicolumn{6}{l}{$\model_{All}$ (7B)}\\
\: w/o \emph{External Executor} & 48.47 & 58.69 & 67.21 & 54.84 & 63.68 \\
\: w \emph{External Executor} & 65.13 & 76.49 & 85.13 & 71.38 & 80.24 \\
\: gains (+) & \textcolor{red}{16.66}  & \textcolor{red}{17.80}  & \textcolor{red}{17.92}  & \textcolor{red}{16.54}  &  \textcolor{red}{16.56} \\
\midrule
\multicolumn{6}{l}{$\model_{All}$ (13B)}\\
\: w/o \emph{External Executor} & 60.05 & 62.60 & 70.73 & 59.95 & 68.61 \\
\: w \emph{External Executor} & 71.75 & 76.79 & 85.05 & 71.86 & 80.50 \\
 \: gains (+) & \textcolor{red}{11.70}  & \textcolor{red}{14.19}  & \textcolor{red}{14.32})  & \textcolor{red}{11.91}  &  \textcolor{red}{11.89} \\
\midrule
\multicolumn{6}{l}{$\model_{All}$ (70B)}\\
\: w/o \emph{External Executor} & 70.10 & 76.61 & 83.55 & 71.74 & 78.99 \\
\: w \emph{External Executor} & \bf 76.81 & \bf 81.42 & \bf 88.49 & \bf 76.55 & \bf 83.90 \\
\: gains (+) & \textcolor{red}{6.71} & \textcolor{red}{4.81}  & \textcolor{red}{4.94}  & \textcolor{red}{4.81}  &  \textcolor{red}{4.91} \\
 \bottomrule
\end{tabular}
 \caption{Effectiveness of the \emph{External Executor} for $\model_{ALL}$ with different sizes. }
    \label{tab:executor}
  \vspace{-2.8em}
\end{table}

\begin{figure}[b]
    \begin{center}
     \vspace{-15pt}
    \includegraphics[scale=0.42]{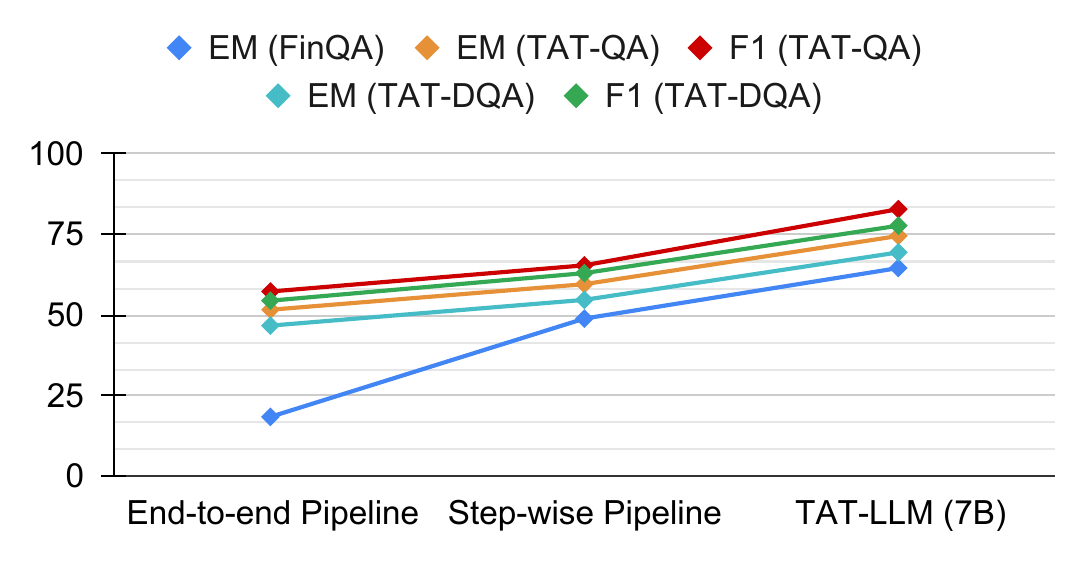}
    \end{center}
    \vspace{-15pt}
    \caption{\label{fig:ablation_study} 
   Comparison of different training strategies.}
\end{figure}

\paragraph{Different Fine-tuning Strategies.} 
\label{module_analysis}
To verify the effect of our fine-tuning strategy, i.e. \emph{Step-wise Pipeline} plus \emph{External Executor}, we compare it with separately applying \emph{End-to-end Pipeline} or \emph{Step-wise Pipeline} on LLaMA 2 (7B) models. 
The results are summarized in \figref{ablation_study}.
Firstly, we can see our \model~(7B) clearly outperforms the two variants following the \emph{End-to-end Pipeline} and the \emph{Step-wise Pipeline} on all tabular and textual QA datasets for both evaluation metrics, demonstrating the effectiveness of our fine-tuning strategy. 
Besides, we find that the \emph{Step-wise Pipeline} also outperforms the \emph{End-to-end Pipeline} in all situations, showing the rationality of incorporating intermediate reasoning steps, which is in line with previous research on chain-of-thoughts reasoning~\cite{wei2022chain}. 
Moreover, it is observed that the \emph{End-to-end Pipeline} achieves notably poor performance on FinQA compared to that on TAT-QA and TAT-DQA. This is probably because FinQA requires more complex discrete reasoning steps than TAT-QA and TAT-DQA, and omitting such steps in fining-tuning would largely degrade model performance. 
Finally, we find a significant performance increase from the \emph{Step-wise Pipeline} to \model. This is due to the incorporation of the reliable \emph{External Executor} which largely guarantees the reasoning accuracy of the answer. In the next section, we perform detailed analysis on the effectiveness of the \emph{External Executor}.

\begin{figure}[t]
    \begin{center}
    \includegraphics[scale=0.43]{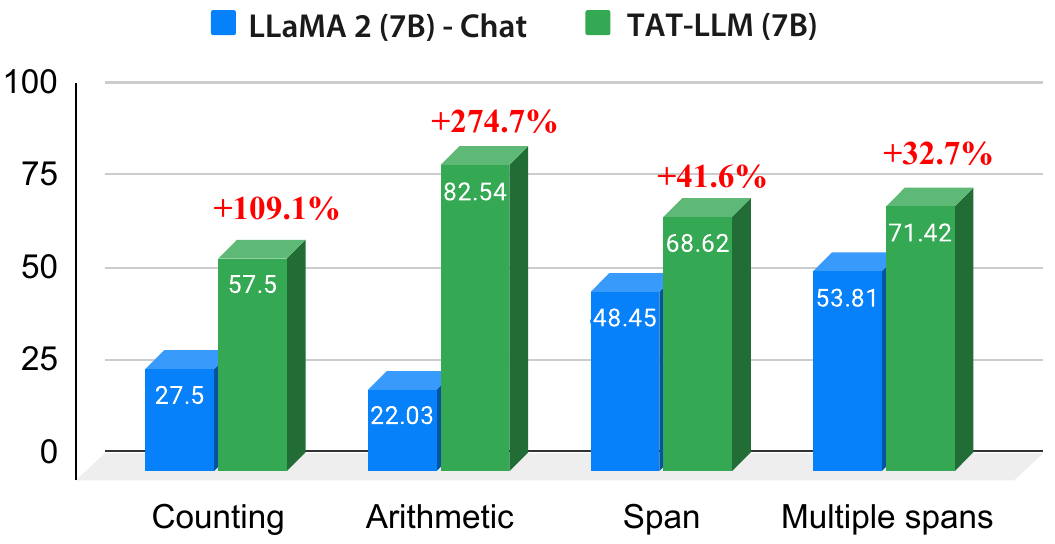}
    \end{center}
    \vspace{-10pt}
   \caption{
   Performance comparison in terms of EM between \model~(7B) and LLaMA 2-Chat (7B) for different question types on TAT-QA. 
   }
   \label{fig:tatqa_question_type}
    \vspace{-0.5cm}
\end{figure}

\paragraph{Effect of \emph{External Executor} w.r.t. Different Model Sizes.}
\label{executor_exp}

Here we analyze the effectiveness of the \emph{External Executor} on models with different sizes. 
According to the experimental results shown in \tabref{executor}, we can see that ablating the \emph{External Executor} significantly degrades the performance across all model sizes on all datasets, causing decreases of larger than $15\%$ on the EM of the three datasets for $\model_{ALL}$ (7B).
This shows that the \emph{External Executor} is essential in ensuring the answer correctness. 
Also, we find that larger models w/o \emph{External Executor} perform better than smaller models w/o \emph{External Executor}, and the gains by the \emph{External Executor} get smaller as the model size increases. The largest $\model_{ALL}$ (70B) w/o \emph{External Executor} only decreases less than 7\% on all three datasets, which is far less than $\model_{ALL}$ (7B).
This is because larger models have stronger discrete reasoning abilities than smaller models and make less mistakes in answer calculation.
Hence, the advantage of ensuring accurate results in discrete reasoning through the \emph{External Executor} appears to be less conspicuous.

\begin{figure}[t]
    \begin{center}
    \includegraphics[scale=0.43]{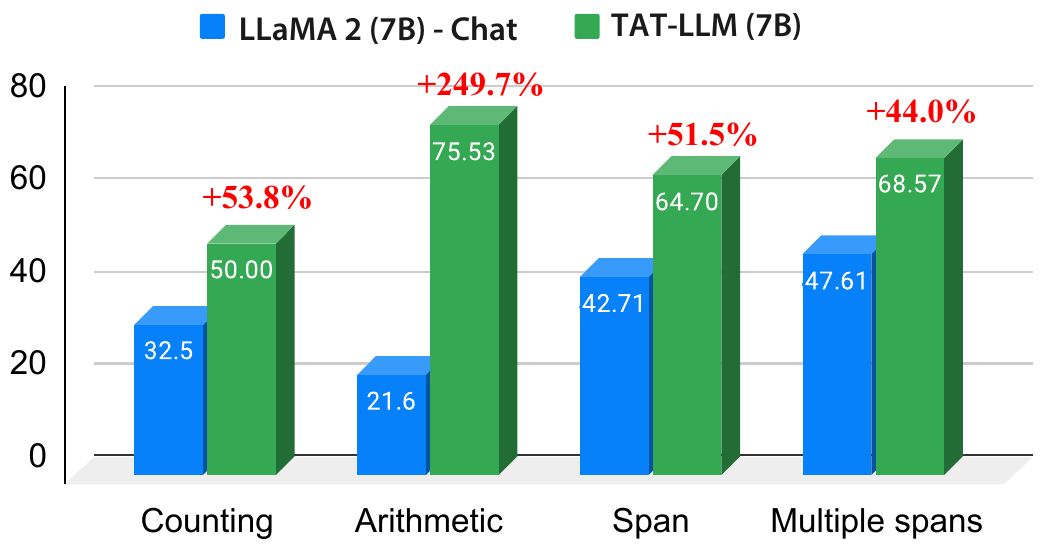}
    \end{center}
    \vspace{-10pt}
    \caption{ 
   Performance comparison in terms of EM between \model~(7B) and LLaMA 2-Chat (7B) for different question types on TAT-DQA.
   }
   \label{fig:tatdqa_question_type}
   \vspace{-1.1em}
\end{figure}

\noindent \textbf{Improvements w.r.t. Different Question Types. }
Here we analyze the improvements achieved by \model~model regarding different question types of TAT-QA and TAT-DQA datasets. 
We compare the performance of \model~(7B) and LLaMA 2-Chat (7B) for each question type. 
The results are summarized in Figure~\ref{fig:tatqa_question_type} and Figure~\ref{fig:tatdqa_question_type} for TAT-QA and TAT-DQA, respectively.
We can observe substantial performance improvements for the \textit{Arithmetic} questions, which are over two-fold, for both datasets.
The larger performance gains are attributed to the effective fine-tuning strategy of \model~ as well as the reliable \emph{External Executor} that ensures accurate execution of the equations and logic rules.
Large performance improvements are also found for the \textit{Counting} questions, namely 109.1\% for TAT-QA and 53.8\% for TAT-DQA. 
This is probably due to executing more correct counting operations. For the \textit{Span} and \textit{Multiple Spans} types, relatively smaller improvements are observed, which are still over 30\%. 
This is potentially because these questions involve simpler or no discrete reasoning, mainly numerical comparison.

\section{Conclusion}
In this work,  we first abstract a \emph{Step-wise Pipeline} for tabular and textual QA based on the previous multi-step methods.
Following this pipeline, we develop \model~model by specializing smaller language models (i.e., 7B) in discrete reasoning over tabular and textual data.
We validate its effectiveness with extensive experiments, showing that our \model~model can outperform both conventional methods and very large-scale LMs like GPT-4 on this task.
Our work well demonstrates that specialized models might be a promising direction towards more advanced models on specific tasks that can compete with human experts.

\section*{Limitations} 
Despite the impressive performance on all three datasets i.e., FinQA~\cite{chen2021finqa}, TAT-QA~\cite{zhu2021tat}, and TAT-DQA~\cite{zhu2022towards}, our \model~model still has much room for further improvement, as shown in error analysis in Appendix \ref{error-analysis}. 
In addition, our \model~model is designed for the documents that contain both tabular and textual data and feature rich numerical values.
This means it may have limited advantages over other kinds of documents like pure textual documents.  
Also, our model may not be directly applied to understand the documents with a large number of pages (e.g., >100 pages) due to the constraint of the maximum input sequence length.

\section*{Ethics Statement}
In this work, we first abstract a \emph{Step-wise Pipeline} for tabular and textual QA.
Then, we propose the \model~model by specializing a smaller language model (i.e., LLaMA 2 7B, 13B and 70B) in discrete reasoning over a hybrid of tabular and textual data.
Our \model~model is developed on open-source tools and datasets to assist human beings in processing and understanding such kind of data.
Thus, we do not anticipate any potential risks or negative ethical issues.

\bibliography{anthology,custom}

\begin{thebibliography}{32}
\expandafter\ifx\csname natexlab\endcsname\relax\def\natexlab#1{#1}\fi

\bibitem[{Brown et~al.(2020)Brown, Mann, Ryder, Subbiah, Kaplan, Dhariwal, Neelakantan, Shyam, Sastry, Askell, Agarwal, Herbert-Voss, Krueger, Henighan, Child, Ramesh, Ziegler, Wu, Winter, Hesse, Chen, Sigler, Litwin, Gray, Chess, Clark, Berner, McCandlish, Radford, Sutskever, and Amodei}]{openai2020gpt3}
Tom Brown, Benjamin Mann, Nick Ryder, Melanie Subbiah, Jared~D Kaplan, Prafulla Dhariwal, Arvind Neelakantan, Pranav Shyam, Girish Sastry, Amanda Askell, Sandhini Agarwal, Ariel Herbert-Voss, Gretchen Krueger, Tom Henighan, Rewon Child, Aditya Ramesh, Daniel Ziegler, Jeffrey Wu, Clemens Winter, Chris Hesse, Mark Chen, Eric Sigler, Mateusz Litwin, Scott Gray, Benjamin Chess, Jack Clark, Christopher Berner, Sam McCandlish, Alec Radford, Ilya Sutskever, and Dario Amodei. 2020.
\newblock Language models are few-shot learners.
\newblock In \emph{Advances in Neural Information Processing Systems}, pages 1877--1901.

\bibitem[{Chen et~al.(2020{\natexlab{a}})Chen, Chang, Schlinger, Wang, and Cohen}]{chen2020open}
Wenhu Chen, Ming-Wei Chang, Eva Schlinger, William~Yang Wang, and William~W Cohen. 2020{\natexlab{a}}.
\newblock Open question answering over tables and text.
\newblock In \emph{International Conference on Learning Representations}.

\bibitem[{Chen et~al.(2020{\natexlab{b}})Chen, Zha, Chen, Xiong, Wang, and Wang}]{chen2020hybridqa}
Wenhu Chen, Hanwen Zha, Zhiyu Chen, Wenhan Xiong, Hong Wang, and William~Yang Wang. 2020{\natexlab{b}}.
\newblock Hybridqa: A dataset of multi-hop question answering over tabular and textual data.
\newblock In \emph{Findings of the Association for Computational Linguistics: EMNLP 2020}, pages 1026--1036.

\bibitem[{Chen et~al.(2021)Chen, Chen, Smiley, Shah, Borova, Langdon, Moussa, Beane, Huang, Routledge, and Wang}]{chen2021finqa}
Zhiyu Chen, Wenhu Chen, Charese Smiley, Sameena Shah, Iana Borova, Dylan Langdon, Reema Moussa, Matt Beane, Ting-Hao Huang, Bryan Routledge, and William~Yang Wang. 2021.
\newblock {F}in{QA}: A dataset of numerical reasoning over financial data.
\newblock In \emph{Proceedings of the 2021 Conference on Empirical Methods in Natural Language Processing}, pages 3697--3711. Association for Computational Linguistics.

\bibitem[{Chiang et~al.(2023)Chiang, Li, Lin, Sheng, Wu, Zhang, Zheng, Zhuang, Zhuang, Gonzalez, Stoica, and Xing}]{vicuna2023}
Wei-Lin Chiang, Zhuohan Li, Zi~Lin, Ying Sheng, Zhanghao Wu, Hao Zhang, Lianmin Zheng, Siyuan Zhuang, Yonghao Zhuang, Joseph~E. Gonzalez, Ion Stoica, and Eric~P. Xing. 2023.
\newblock \href {https://lmsys.org/blog/2023-03-30-vicuna/} {Vicuna: An open-source chatbot impressing gpt-4 with 90\%* chatgpt quality}.

\bibitem[{Cobbe et~al.(2021)Cobbe, Kosaraju, Bavarian, Chen, Jun, Kaiser, Plappert, Tworek, Hilton, Nakano, Hesse, and Schulman}]{cobbe2021gsm8k}
Karl Cobbe, Vineet Kosaraju, Mohammad Bavarian, Mark Chen, Heewoo Jun, Lukasz Kaiser, Matthias Plappert, Jerry Tworek, Jacob Hilton, Reiichiro Nakano, Christopher Hesse, and John Schulman. 2021.
\newblock Training verifiers to solve math word problems.
\newblock \emph{arXiv preprint arXiv:2110.14168}.

\bibitem[{Fu et~al.(2023)Fu, Peng, Ou, Sabharwal, and Khot}]{fu2023specializing}
Yao Fu, Hao Peng, Litu Ou, Ashish Sabharwal, and Tushar Khot. 2023.
\newblock Specializing smaller language models towards multi-step reasoning.
\newblock In \emph{Proceedings of the 40th International Conference on Machine Learning}, volume 202 of \emph{Proceedings of Machine Learning Research}, pages 10421--10430. PMLR.

\bibitem[{Hu et~al.(2022)Hu, yelong shen, Wallis, Allen-Zhu, Li, Wang, Wang, and Chen}]{hu2022lora}
Edward~J Hu, yelong shen, Phillip Wallis, Zeyuan Allen-Zhu, Yuanzhi Li, Shean Wang, Lu~Wang, and Weizhu Chen. 2022.
\newblock \href {https://openreview.net/forum?id=nZeVKeeFYf9} {Lo{RA}: Low-rank adaptation of large language models}.
\newblock In \emph{International Conference on Learning Representations}.

\bibitem[{Lei et~al.(2022)Lei, He, Li, Zhao, and Liu}]{lei2022answering}
Fangyu Lei, Shizhu He, Xiang Li, Jun Zhao, and Kang Liu. 2022.
\newblock \href {https://aclanthology.org/2022.coling-1.118} {Answering numerical reasoning questions in table-text hybrid contents with graph-based encoder and tree-based decoder}.
\newblock In \emph{Proceedings of the 29th International Conference on Computational Linguistics}, pages 1379--1390. International Committee on Computational Linguistics.

\bibitem[{Li et~al.(2022)Li, Feng, Zhang, He, Zhu, and Chua}]{li2022learning}
Moxin Li, Fuli Feng, Hanwang Zhang, Xiangnan He, Fengbin Zhu, and Tat-Seng Chua. 2022.
\newblock Learning to imagine: Integrating counterfactual thinking in neural discrete reasoning.
\newblock In \emph{Proceedings of the 60th Annual Meeting of the Association for Computational Linguistics (Volume 1: Long Papers)}, pages 57--69. Association for Computational Linguistics.

\bibitem[{Li et~al.(2021)Li, Sun, and Cheng}]{li2021tsqa}
Xiao Li, Yawei Sun, and Gong Cheng. 2021.
\newblock Tsqa: tabular scenario based question answering.
\newblock In \emph{Proceedings of the AAAI Conference on Artificial Intelligence}, volume~35, pages 13297--13305.

\bibitem[{Li et~al.(2023)Li, Zhu, Liu, Ju, Qu, and Cheng}]{Li2023DyRRen}
Xiao Li, Yin Zhu, Sichen Liu, Jiangzhou Ju, Yuzhong Qu, and Gong Cheng. 2023.
\newblock Dyrren: A dynamic retriever-reranker-generator model for numerical reasoning over tabular and textual data.
\newblock AAAI Press.

\bibitem[{Luo et~al.(2023)Luo, Sun, Xu, Zhao, Lou, Tao, Geng, Lin, Chen, and Zhang}]{luo2023wizardmath}
Haipeng Luo, Qingfeng Sun, Can Xu, Pu~Zhao, Jianguang Lou, Chongyang Tao, Xiubo Geng, Qingwei Lin, Shifeng Chen, and Dongmei Zhang. 2023.
\newblock \href {http://arxiv.org/abs/2308.09583} {Wizardmath: Empowering mathematical reasoning for large language models via reinforced evol-instruct}.

\bibitem[{OpenAI(2023)}]{openai2023gpt4}
OpenAI. 2023.
\newblock \href {http://arxiv.org/abs/2303.08774} {Gpt-4 technical report}.

\bibitem[{Ouyang et~al.(2022)Ouyang, Wu, Jiang, Almeida, Wainwright, Mishkin, Zhang, Agarwal, Slama, Ray, Schulman, Hilton, Kelton, Miller, Simens, Askell, Welinder, Christiano, Leike, and Lowe}]{ouyang2023following}
Long Ouyang, Jeffrey Wu, Xu~Jiang, Diogo Almeida, Carroll Wainwright, Pamela Mishkin, Chong Zhang, Sandhini Agarwal, Katarina Slama, Alex Ray, John Schulman, Jacob Hilton, Fraser Kelton, Luke Miller, Maddie Simens, Amanda Askell, Peter Welinder, Paul~F Christiano, Jan Leike, and Ryan Lowe. 2022.
\newblock Training language models to follow instructions with human feedback.
\newblock In \emph{Advances in Neural Information Processing Systems}, volume~35, pages 27730--27744. Curran Associates, Inc.

\bibitem[{Ran et~al.(2019)Ran, Lin, Li, Zhou, and Liu}]{ran2019numnet}
Qiu Ran, Yankai Lin, Peng Li, Jie Zhou, and Zhiyuan Liu. 2019.
\newblock {N}um{N}et: Machine reading comprehension with numerical reasoning.
\newblock In \emph{EMNLP-IJCNLP}, pages 2474--2484.

\bibitem[{Rogers et~al.(2023)Rogers, Gardner, and Augenstein}]{10.1145/3560260}
Anna Rogers, Matt Gardner, and Isabelle Augenstein. 2023.
\newblock \href {https://doi.org/10.1145/3560260} {Qa dataset explosion: A taxonomy of nlp resources for question answering and reading comprehension}.
\newblock \emph{ACM Comput. Surv.}, 55(10).

\bibitem[{Taori et~al.(2023)Taori, Gulrajani, Zhang, Dubois, Li, Guestrin, Liang, and Hashimoto}]{alpaca}
Rohan Taori, Ishaan Gulrajani, Tianyi Zhang, Yann Dubois, Xuechen Li, Carlos Guestrin, Percy Liang, and Tatsunori~B. Hashimoto. 2023.
\newblock Stanford alpaca: An instruction-following llama model.
\newblock \url{https://github.com/tatsu-lab/stanford_alpaca}.

\bibitem[{Touvron et~al.(2023{\natexlab{a}})Touvron, Lavril, Izacard, Martinet, Lachaux, Lacroix, Rozière, Goyal, Hambro, Azhar, Rodriguez, Joulin, Grave, and Lample}]{touvron2023llama}
Hugo Touvron, Thibaut Lavril, Gautier Izacard, Xavier Martinet, Marie-Anne Lachaux, Timothée Lacroix, Baptiste Rozière, Naman Goyal, Eric Hambro, Faisal Azhar, Aurelien Rodriguez, Armand Joulin, Edouard Grave, and Guillaume Lample. 2023{\natexlab{a}}.
\newblock \href {http://arxiv.org/abs/2302.13971} {Llama: Open and efficient foundation language models}.

\bibitem[{Touvron et~al.(2023{\natexlab{b}})Touvron, Martin, Stone, Albert, Almahairi, Babaei, Bashlykov, Batra, Bhargava, Bhosale, Bikel, Blecher, Ferrer, Chen, Cucurull, Esiobu, Fernandes, Fu, Fu, Fuller, Gao, Goswami, Goyal, Hartshorn, Hosseini, Hou, Inan, Kardas, Kerkez, Khabsa, Kloumann, Korenev, Koura, Lachaux, Lavril, Lee, Liskovich, Lu, Mao, Martinet, Mihaylov, Mishra, Molybog, Nie, Poulton, Reizenstein, Rungta, Saladi, Schelten, Silva, Smith, Subramanian, Tan, Tang, Taylor, Williams, Kuan, Xu, Yan, Zarov, Zhang, Fan, Kambadur, Narang, Rodriguez, Stojnic, Edunov, and Scialom}]{touvron2023llama2}
Hugo Touvron, Louis Martin, Kevin Stone, Peter Albert, Amjad Almahairi, Yasmine Babaei, Nikolay Bashlykov, Soumya Batra, Prajjwal Bhargava, Shruti Bhosale, Dan Bikel, Lukas Blecher, Cristian~Canton Ferrer, Moya Chen, Guillem Cucurull, David Esiobu, Jude Fernandes, Jeremy Fu, Wenyin Fu, Brian Fuller, Cynthia Gao, Vedanuj Goswami, Naman Goyal, Anthony Hartshorn, Saghar Hosseini, Rui Hou, Hakan Inan, Marcin Kardas, Viktor Kerkez, Madian Khabsa, Isabel Kloumann, Artem Korenev, Punit~Singh Koura, Marie-Anne Lachaux, Thibaut Lavril, Jenya Lee, Diana Liskovich, Yinghai Lu, Yuning Mao, Xavier Martinet, Todor Mihaylov, Pushkar Mishra, Igor Molybog, Yixin Nie, Andrew Poulton, Jeremy Reizenstein, Rashi Rungta, Kalyan Saladi, Alan Schelten, Ruan Silva, Eric~Michael Smith, Ranjan Subramanian, Xiaoqing~Ellen Tan, Binh Tang, Ross Taylor, Adina Williams, Jian~Xiang Kuan, Puxin Xu, Zheng Yan, Iliyan Zarov, Yuchen Zhang, Angela Fan, Melanie Kambadur, Sharan Narang, Aurelien Rodriguez, Robert Stojnic, Sergey Edunov, and Thomas
  Scialom. 2023{\natexlab{b}}.
\newblock \href {http://arxiv.org/abs/2307.09288} {Llama 2: Open foundation and fine-tuned chat models}.

\bibitem[{Wei et~al.(2022{\natexlab{a}})Wei, Bosma, Zhao, Guu, Yu, Lester, Du, Dai, and Le}]{wei2022finetuned}
Jason Wei, Maarten Bosma, Vincent Zhao, Kelvin Guu, Adams~Wei Yu, Brian Lester, Nan Du, Andrew~M. Dai, and Quoc~V Le. 2022{\natexlab{a}}.
\newblock \href {https://openreview.net/forum?id=gEZrGCozdqR} {Finetuned language models are zero-shot learners}.
\newblock In \emph{International Conference on Learning Representations}.

\bibitem[{Wei et~al.(2022{\natexlab{b}})Wei, Tay, Bommasani, Raffel, Zoph, Borgeaud, Yogatama, Bosma, Zhou, Metzler, Chi, Hashimoto, Vinyals, Liang, Dean, and Fedus}]{wei2022emergent}
Jason Wei, Yi~Tay, Rishi Bommasani, Colin Raffel, Barret Zoph, Sebastian Borgeaud, Dani Yogatama, Maarten Bosma, Denny Zhou, Donald Metzler, Ed~H. Chi, Tatsunori Hashimoto, Oriol Vinyals, Percy Liang, Jeff Dean, and William Fedus. 2022{\natexlab{b}}.
\newblock Emergent abilities of large language models.
\newblock \emph{Transactions on Machine Learning Research}.

\bibitem[{Wei et~al.(2022{\natexlab{c}})Wei, Wang, Schuurmans, Bosma, brian ichter, Xia, Chi, Le, and Zhou}]{wei2022chain}
Jason Wei, Xuezhi Wang, Dale Schuurmans, Maarten Bosma, brian ichter, Fei Xia, Ed~H. Chi, Quoc~V Le, and Denny Zhou. 2022{\natexlab{c}}.
\newblock \href {https://openreview.net/forum?id=_VjQlMeSB_J} {Chain of thought prompting elicits reasoning in large language models}.
\newblock In \emph{Advances in Neural Information Processing Systems}.

\bibitem[{Wei et~al.(2023)Wei, Lei, Zhang, Zhao, and Liu}]{wei2023multiview}
Yifan Wei, Fangyu Lei, Yuanzhe Zhang, Jun Zhao, and Kang Liu. 2023.
\newblock \href {http://arxiv.org/abs/2305.03458} {Multi-view graph representation learning for answering hybrid numerical reasoning question}.

\bibitem[{Yue et~al.(2023)Yue, Qu, Zhang, Fu, Huang, Sun, Su, and Chen}]{yue2023mammoth}
Xiang Yue, Xingwei Qu, Ge~Zhang, Yao Fu, Wenhao Huang, Huan Sun, Yu~Su, and Wenhu Chen. 2023.
\newblock \href {http://arxiv.org/abs/2309.05653} {Mammoth: Building math generalist models through hybrid instruction tuning}.

\bibitem[{Zhao et~al.(2022)Zhao, Li, Li, and Zhang}]{zhao2022multihiertt}
Yilun Zhao, Yunxiang Li, Chenying Li, and Rui Zhang. 2022.
\newblock {M}ulti{H}iertt: Numerical reasoning over multi hierarchical tabular and textual data.
\newblock In \emph{Proceedings of the 60th Annual Meeting of the Association for Computational Linguistics (Volume 1: Long Papers)}, pages 6588--6600. Association for Computational Linguistics.

\bibitem[{Zhou et~al.(2023)Zhou, Sch{\"a}rli, Hou, Wei, Scales, Wang, Schuurmans, Cui, Bousquet, Le, and Chi}]{zhou2023leasttomost}
Denny Zhou, Nathanael Sch{\"a}rli, Le~Hou, Jason Wei, Nathan Scales, Xuezhi Wang, Dale Schuurmans, Claire Cui, Olivier Bousquet, Quoc~V Le, and Ed~H. Chi. 2023.
\newblock \href {https://openreview.net/forum?id=WZH7099tgfM} {Least-to-most prompting enables complex reasoning in large language models}.
\newblock In \emph{The Eleventh International Conference on Learning Representations}.

\bibitem[{Zhou et~al.(2022)Zhou, Bao, Duan, Wu, He, and Zhao}]{zhou2022unirpg}
Yongwei Zhou, Junwei Bao, Chaoqun Duan, Youzheng Wu, Xiaodong He, and Tiejun Zhao. 2022.
\newblock Unirpg: Unified discrete reasoning over table and text as program generation.
\newblock \emph{arXiv preprint arXiv:2210.08249}.

\bibitem[{Zhu et~al.(2022)Zhu, Lei, Feng, Wang, Zhang, and Chua}]{zhu2022towards}
Fengbin Zhu, Wenqiang Lei, Fuli Feng, Chao Wang, Haozhou Zhang, and Tat-Seng Chua. 2022.
\newblock Towards complex document understanding by discrete reasoning.
\newblock In \emph{Proceedings of the 30th ACM International Conference on Multimedia}, pages 4857--4866.

\bibitem[{Zhu et~al.(2021)Zhu, Lei, Huang, Wang, Zhang, Lv, Feng, and Chua}]{zhu2021tat}
Fengbin Zhu, Wenqiang Lei, Youcheng Huang, Chao Wang, Shuo Zhang, Jiancheng Lv, Fuli Feng, and Tat-Seng Chua. 2021.
\newblock {TAT}-{QA}: A question answering benchmark on a hybrid of tabular and textual content in finance.
\newblock In \emph{Proceedings of the 59th Annual Meeting of the Association for Computational Linguistics and the 11th International Joint Conference on Natural Language Processing (Volume 1: Long Papers)}, pages 3277--3287. Association for Computational Linguistics.

\bibitem[{Zhu et~al.(2023)Zhu, Li, Xiao, Feng, Wang, and Chua}]{Zhu2023SoarGraph}
Fengbin Zhu, Moxin Li, Junbin Xiao, Fuli Feng, Chao Wang, and Tat~Seng Chua. 2023.
\newblock Soargraph: Numerical reasoning over financial table-text data via semantic-oriented hierarchical graphs.
\newblock In \emph{Companion Proceedings of the ACM Web Conference 2023}, page 1236–1244. Association for Computing Machinery.

\bibitem[{Ziqi and Lu(2023)}]{ziqilu2023tabcot}
Jin Ziqi and Wei Lu. 2023.
\newblock Tab-{C}o{T}: Zero-shot tabular chain of thought.
\newblock In \emph{Findings of the Association for Computational Linguistics: ACL 2023}, pages 10259--10277. Association for Computational Linguistics.

\end{thebibliography}
\bibliographystyle{acl_natbib}

\appendix

\label{sec:appendix}
\clearpage
\appendix
\setcounter{table}{0}
\renewcommand{\tabref}[1]{Table~A\ref{tab:#1}}

\section{Implementation Details}
\label{Implementation Details}
Since the question type and the scale of the answer have been annotated in both TAT-QA and TAT-DQA datasets, we add two more steps in the instruction part of the input instance, in addition to the three steps in the proposed \emph{Step-wise Pipeline} shown in \figref{qa_system} c). 
One step is Question Type Predictor, 
with which we limit the model to select one value from the following question types: \textit{Span}, \textit{Multiple Spans}, \textit{Counting}, and \textit{Arithmetic}.
For \textit{Span} and \textit{Multiple Spans} questions, we set the output of the \textit{Reasoner} to ``N.A.''. 
The other step is Scale Predictor, with which we restrict the model to choose one from the following values: \textit{Thousand}, \textit{Million}, \textit{Billion}, \textit{Percent}  and \textit{None}.
For all the templates used to construct training instances of our ~\model~model, please refer to Appendix~\ref{appe:template_finetuning}. 


We train our \model~model on one NVIDIA DGX-A100 with eight A100 GPUs.
The quantization is 8 bit.
We use Adam optimizer with a learning rate of $3e-4$ and warmup over the first $3\%$ steps to train. 
The maximum sequence length is $4,096$ and the maximum number of epochs is set to $3$. 
The batch size is set to $4$ and the gradient accumulation step is $10$. 

For comparing with fine-tuned models, we take the results from their original papers respectively. 
For each LLM, we utilize three different templates to perform zero-shot inference to obtain prediction results. 
We select the best result as the reported result in Section~\ref{main_result}.
Please refer to Appendix~\ref{appe:template_zero} for details of the templates we use for zero-shot inference.
We utilize the latest version\footnote{Date: Sep 2023} 
GPT3.5-Turbo~\cite{ouyang2023following} and GPT-4~\cite{openai2023gpt4} via OpenAI APIs.
We set the temperature as 0, top p as 1.0, max token as $1,000$, and other parameters as default. 
We obtain the official trained checkpoint of Vicuna~\cite{vicuna2023}, LLaMA 2-Chat~\cite{touvron2023llama2}, MAmmoTH~\cite{yue2023mammoth} and WizardMath~\cite{luo2023wizardmath} from Hugginface.
The inference is done on one NVIDIA DGX-A100 with eight A100 GPUs. 
The parameters 
\texttt{num\_beam} and \texttt{do\_sample} are 1 and false respectively.

\begin{table}[t]
\centering
\footnotesize
\begin{tabular} 
{p{0.15\linewidth}|p{0.15\linewidth}|p{0.56\linewidth}}
\toprule
\bf  Step & \bf Error(\%) & \bf Example \\
\toprule
\multirow{12}{*}{\bf Extractor} &  
\multirow{3}{*}{\shortstack[l]{Wrong \\ Evidence \\(48\%)}}  
& \textbf{Q}: What is the percentage change in cash flow hedges in 2011 compared to 2010? \\
& & \textbf{G}: 153.7, \textcolor{blue}{139.9} \\
& & \textbf{P}: 153.7, \textcolor{red}{375.0} \\
\cmidrule{2-3}
& \multirow{3}{*}{\shortstack[l]{Missed \\ Evidence \\ (15\%)}} 
& \textbf{Q}: What was the total impairment costs recorded from 2003 to 2005 in millions? \\
& & \textbf{G}: 0.6, 0.5, \textcolor{blue}{4.7} \\
& & \textbf{P}: 0.6, 0.5 \\
\cmidrule{2-3}
& \multirow{3}{*}{\shortstack[l]{Redundant \\ Evidence \\ (8\%)}} 
& \textbf{Q}: What was the average number of shares issued to employees from 2013 to 2015? \\
& & \textbf{G}: 439000, 411636 \\
& & \textbf{P}: 439000, 411636, \textcolor{red}{556000} \\
\midrule
\multirow{11}{*}{\bf Reasoner} &  
\multirow{3}{*}{\shortstack[l]{Wrong \\ Operators \\ (19\%)}}  
& \textbf{Q}: what was the percent of the change in the stock price performance for hum from 2010 to 2011? \\
& & \textbf{G}: \textcolor{blue}{(201 - 125) / 125}  \\
& & \textbf{P}: \textcolor{red}{201 - 125} \\
\cmidrule{2-3}
& \multirow{3}{*}{\shortstack[l]{Wrong \\ Values  \\ (10\%)}} 
& \textbf{Q}: What was the average cash flow from 2004 to 2006? \\
& & \textbf{G}: (\textcolor{blue}{950.4} + 957.4 + 769.1) / 3 \\
& & \textbf{P}: (\textcolor{red}{957.4} + 957.4 + 769.1) / 3\\
\bottomrule
\end{tabular} 
\caption{
Examples of errors on the test set of FinQA and corresponding percentages in each step. 
Q, G and P denote question, ground truth and prediction.}
\label{tab:error_cases} 
\vspace{-2em}
\end{table}

\section{Error Analysis}
\label{error-analysis}
To further diagnose our \model~model, we randomly sample $100$ error instances of \model~(7B) from the test set of FinQA and analyze the reasons.
Since we adopt the \emph{External Executor} for reliable execution, the errors only occur to the \textit{Extractor} and \textit{Reasoner}.
As shown in \tabref{error_cases}, we list all kinds of errors with examples and their corresponding percentages. 
We can observe that most of the errors come from the \emph{Extractor}, where the \emph{Wrong Evidence} takes $48\%$ of the total errors. 
This shows that our \model~(7B) model still faces challenges in precisely interpreting the meaning of the values, sometimes due to the unique terminology. 
Training the model with more tabular and textual data in this domain might be a possible approach to enhancing its evidence extraction ability. 
Additionally, the \emph{Wrong Operator} and the \emph{Wrong Values} may be partially caused by the randomness of the LLM as the generation length increases, which fails to generate the full equation or wrongly copies the extracted evidence.

\section{Templates for Fine-tuning } \label{appe:template_finetuning}

To prepare data for training our \model~model, the below templates are applied to construct the training instances following the \emph{Step-wise Pipeline} on the three tabular and textual QA datasets, i.e., \tabref{template_sft_finqa} for FinQA, \tabref{template_sft_tatqa} for TAT-QA and \tabref{template_sft_tatdqa} for TAT-DQA, respectively.

For comparison, we also present the templates we used to prepare the training instances following the \emph{End-to-end Pipeline} in \tabref{template_sft_finqa_sd} for FinQA, \tabref{template_sft_tatqa_sd} for TAT-QA, and \tabref{template_sft_tatdqa_sd} for TAT-DQA.

\begin{table*}[h]
    \centering
    \small
    \begin{tabular}{p{15cm}}
    \toprule
       Below is an instruction that describes a question answering task in the finance domain, paired with an input table and its relevant text that provide further context. The given question is relevant to the table and text. Generate an appropriate answer to the given question. \\
        \\
        \#\#\# Instruction:
        
        Given a table and a list of texts in the following, what is the answer to the question? 
        Please complete the task in three steps:
        
        1. In the first step, extract the relevant numerical values from the provided table or texts. Store these in the variable `\{evidence\}`. If there are multiple values, separate them using the '\#' symbol.
        
        2. In the second step, generate an equation using the extracted numerical values. Store this equation in the variable `\{equation\}`.
        
        3. In the third step, calculate the answer based on the equation and store it in the variable `\{answer\}`.\\
        
        Please organize the results in the following table:
        
        | step | output  |
        
        | 1 | \{evidence\} |
        
        | 2 | \{equation\} |
        
        | 3 | \{answer\}   |\\

        Finally, present the calculated answer in the format: "The answer is: \{answer\}" \\
        \\
        \#\#\# Table 
        
        \{table\}\\
               \\
        \#\#\# Text
        
        \{text\}\\
               \\
        \#\#\# Question 
        
        \{question\}\\
               \\
        \#\#\# Response 
        
        |step | output|
        
        |1 | \{gold\_evidence\} |
        
        |2 | \{gold\_equation\} |
        
        |3 | \{gold\_answer\} |\\
        
        The answer is: \{gold\_answer\} \\
    \bottomrule
    \end{tabular}
    \caption{
    The template for constructing training instances following \emph{Step-wise Pipeline} with FinQA dataset.}
    \label{tab:template_sft_finqa}
\end{table*}

\begin{table*}[h]
    \centering
    \small
    \begin{tabular}{p{15cm}}
    \toprule

Below is an instruction that describes a question answering task in the finance domain, paired with an input table and its relevant text that provide further context. The given question is relevant to the table and text. Generate an appropriate answer to the given question.\\
\\
\#\#\# Instruction

Given a table and a list of texts in the following, answer the question posed using the following five-step process: \\

1. Step 1: Predict the type of question being asked. Store this prediction in the variable `\{question\_type\}`. The value of `\{question\_type\}` can be one of the following:`Single span`, `Multiple spans`, `Count`, or `Arithmetic`.

2. Step 2: Extract the relevant strings or numerical values from the provided table or texts. Store these pieces of evidence in the variable `\{evidence\}`. If there are multiple pieces of evidence, separate them using the '\#' symbol.

3. Step 3: if the `\{question\_type\}` is `Arithmetic`, formulate an equation using values stored in `\{evidence\}`. Store this equation in the variable `\{equation\}`. For all other question types, set the value of \{equation\} to 'N.A.'.

4. Step 4: Predict or calculate the answer based on the question type, evidence and equation. Store it in the variable `\{answer\}`. If there are multiple values, separate them using the '\#' symbol.

5. Step 5: If the value of the `\{answer\}` is numerical, predict its scale and store it in a variable named `\{scale\}`. The value of `\{scale\}` can be one of the following: `none`, `percent`, `thousand`, `million`, or `billion`. For non-numerical values, set the value of `\{scale\}` to 'none'.\\

Please organize the results in the following table:

| step | output |

| 1 | \{question\_type\} |

| 2 | \{evidence\} |

| 3 | \{equation\} |

| 4 | \{answer\} |

| 5 | \{scale\} |\\

Finally, present the final answer in the format: "The answer is: \{answer\} \#\#\#\# and its corresponding scale is: \{scale\}"\\
\\
\#\#\# Table

\{table\}\\
\\
\#\#\# Text

\{text\}\\
\\
\#\#\# Question 

\{question\}\\
\\
\#\#\# Response 

| step | output |

| 1 | \{gold\_question\_type\} |

| 2 | \{gold\_evidence\} |

| 3 | \{gold\_equation\} |

| 4 | \{gold\_answer\} |

| 5 | \{gold\_scale\} |\\

The answer is: \{gold\_answer\} \#\#\#\# and its corresponding scale is: \{gold\_scale\} \\

    \bottomrule
    \end{tabular}
    \caption{
    The template for constructing training instances following \emph{Step-wise Pipeline} with TAT-QA dataset.}
    \label{tab:template_sft_tatqa}
\end{table*}

\begin{table*}[h]
    \centering
    \small
    \begin{tabular}{p{15cm}}
    \toprule
      Below is an instruction that describes a question answering task in the finance domain, paired with an input document that has one or multiple pages that provide further context. The given question is relevant to the document. Generate an appropriate answer to the given question. \\
\\
\#\#\# Instruction

Given a document that has one or multiple pages in the following, answer the question posed using the following five-step process: \\

1. Step 1: Predict the type of question being asked. Store this prediction in the variable `\{question\_type\}`. The value of `\{question\_type\}` can be one of the following:`Single span`, `Multiple spans`, `Count`, or `Arithmetic`.

2. Step 2: Extract the relevant strings or numerical values from the provided document. Store these pieces of evidence in the variable `\{evidence\}`. If there are multiple pieces of evidence, separate them using the '\#' symbol.

3. Step 3: if the `\{question\_type\}` is `Arithmetic`, formulate an equation using values stored in `\{evidence\}`. Store this equation in the variable `\{equation\}`. For all other question types, set the value of \{equation\} to 'N.A.'.

4. Step 4: Predict or calculate the answer based on the question type, evidence and equation. Store it in the variable `\{answer\}`. If there are multiple values, separate them using the '\#' symbol.

5. Step 5: If the value of the `\{answer\}` is numerical, predict its scale and store it in a variable named `\{scale\}`. The value of `\{scale\}` can be one of the following: `none`, `percent`, `thousand`, `million`, or `billion`. For non-numerical values, set the value of `\{scale\}` to 'none'. \\

Please organize the results in the following table:

| step | output |

| 1 | \{question\_type\} |

| 2 | \{evidence\} |

| 3 | \{equation\} |

| 4 | \{answer\} |

| 5 | \{scale\} |\\

Finally, present the final answer in the format: "The answer is: \{answer\} \#\#\#\# and its corresponding scale is: \{scale\}"\\
\\
\#\#\# Text 

\{pages\}\\
\\
\#\#\# Question 

\{question\}\\
\\
\#\#\# Response 

| step | output |

| 1 | \{gold\_question\_type\} |

| 2 | \{gold\_evidence\} |

| 3 | \{gold\_equation\} |

| 4 | \{gold\_answer\} |

| 5 | \{gold\_scale\} |\\

The answer is: \{gold\_answer\} \#\#\#\# and its corresponding scale is: \{gold\_scale\} \\
    \bottomrule
    \end{tabular}
    \caption{
    The template for constructing training instances following \emph{Step-wise Pipeline} with TAT-DQA dataset.}
    \label{tab:template_sft_tatdqa}
\end{table*}

\begin{table*}[h]
    \centering
    \small
    \begin{tabular}{p{15cm}}
    \toprule
Below is an instruction that describes a question answering task in the finance domain, paired with an input table and its relevant text that provide further context. The given question is relevant to the table and text. Generate an appropriate answer to the given question.\\
\\
\#\#\# Instruction\\
Given a table and a list of texts in the following, what is the answer to the question? 
Please output the answer in the format of "The answer is:".

\\
\#\#\# Table \\
\{table\}

\\
\#\#\# Text\\
\{text\}

\\
\#\#\# Question \\
\{question\}

\\
\#\#\# Response\\
The answer is: \{answer\}\\
    \bottomrule
    \end{tabular}
    \caption{
    The template for constructing training instances following \emph{End-to-end Pipeline} with FinQA dataset}
    \label{tab:template_sft_finqa_sd}
\end{table*}

\begin{table*}[h]
    \centering
    \small
    \begin{tabular}{p{15cm}}
    \toprule
Below is an instruction that describes a question answering task in the finance domain, paired with an input table and its relevant text that provide further context. The given question is relevant to the table and text. Generate an appropriate answer to the given question.\\

\\
\#\#\# Instruction\\
Given a table and a list of texts in the following, what is the answer to the question? 
Please predict the answer and store it in a variable named `\{answer\}`. If there are multiple values, separate them using the '\#' symbol.
If the value of the `\{answer\}` is numerical, predict its scale and store it in a variable named `\{scale\}`. 
The value of `\{scale\}` can be one of the following: `none`, `percent`, `thousand`, `million`, or `billion`. For non-numerical values, set the value of `\{scale\}` to 'none'.
Finally, present the final answer in the format of "The answer is: \{answer\} \#\#\#\# and its corresponding scale is: \{scale\}"\\

\\
\#\#\# Table 

\{table\}

\\
\#\#\# Text 

\{text\}

\\
\#\#\# Question \\

\{question\}

\\
\#\#\# Response 

The answer is: \{gold\_answer\} \#\#\#\# and its corresponding scale is: \{gold\_scale\} \\
    \bottomrule
    \end{tabular}
    \caption{
    The template for constructing training instances following \emph{End-to-end Pipeline} with TAT-QA dataset}
    \label{tab:template_sft_tatqa_sd}
\end{table*}

\begin{table*}[h]
    \centering
    \small
    \begin{tabular}{p{15cm}}
    \toprule
Below is an instruction that describes a question answering task in the finance domain, paired with an input document that has one or multiple pages that provide further context. The given question is relevant to the document. Generate an appropriate answer to the given question.\\

\\
\#\#\# Instruction\\

Given a document that has one or multiple pages in the following, what is the answer to the question? 
Please predict the answer and store it in a variable named `\{answer\}`. If there are multiple values, separate them using the '\#' symbol.
If the value of the `\{answer\}` is numerical, predict its scale and store it in a variable named `\{scale\}`. 
The value of `\{scale\}` can be one of the following: `none`, `percent`, `thousand`, `million`, or `billion`. For non-numerical values, set the value of `\{scale\}` to 'none'. 
Finally, present the final answer in the format of "The answer is: \{answer\} \#\#\#\# and its corresponding scale is: \{scale\}"\\

\\
\#\#\# Document 

\{pages\}

\\
\#\#\# Question \\

\{question\}

\\
\#\#\# Response 

The answer is: \{gold\_answer\} \#\#\#\# and its corresponding scale is: \{gold\_scale\} \\
    \bottomrule
    \end{tabular}
    \caption{
    The template for constructing training instances following \emph{End-to-end Pipeline} with TAT-DQA dataset}
    \label{tab:template_sft_tatdqa_sd}
\end{table*}

\section{Templates for Zero-shot Inference } \label{appe:template_zero}

For zero-shot inference with the baseline LLMs such as GPT-3.5-Turbo~\cite{ouyang2023following} and GPT-4~\cite{openai2023gpt4}, we utilize three different 
templates to build the instances and feed them to the LLMs to get the predictions.
We report the best results in Section \ref{main_result}. 
The following are details of the templates we used.
1) \emph{End-to-end Pipeline} template, i.e., \tabref{zero_finqa} for FinQA and TAT-QA, and \tabref{zero_tatdqa} for TAT-DQA; 
2) \emph{Chain-of-Thought template}, which adds "Let's complete this task step by step" to the \emph{End-to-end Pipeline} template; 
3) \emph{Step-wise Pipeline} template that is similar to the templates for fine-tuning, with the part of ``Response'' removed.

\begin{table*}[h]
    \centering
    \small
    \begin{tabular}{p{15cm}}
    \toprule
Below is an instruction that describes a question answering task in the finance domain, paired with an input table and its relevant text that provide further context. The given question is relevant to the table and text. Generate an appropriate answer to the given question. \\

\\
\#\#\# Instruction

Given a table and a list of texts in the following, what is the answer to the question? \textcolor{blue}{(Let's complete this task step by step.)} Please output the answer in the format of "The answer is:". \\

\\
\#\#\# Table 

\{table\} \\

\\
\#\#\# Text

\{text\} \\

\\
\#\#\# Question 

\{question\} 

\\
\#\#\# Response
\\
    \bottomrule
    \end{tabular}
    \caption{
    The template for zero-shot inference using the baseline LLMs on FinQA and TAT-QA datasets.}
    \label{tab:zero_finqa}
\end{table*}

\begin{table*}[h]
    \centering
    \small
    \begin{tabular}{p{15cm}}
    \toprule

Below is an instruction that describes a question answering task in the finance domain, paired with an input document that has one or multiple pages that provide further context. The given question is relevant to the document. Generate an appropriate answer to the given question.\\
\\
\#\#\# Instruction

Given a document with one or multiple pages in the following, what is the answer to the question? 
\textcolor{blue}{(Let's complete this task step by step.)} Please output the answer in the format of "The answer is:".\\
\\
\#\#\# Text 

\{pages\}\\
\\
\#\#\# Question 

\{question\}\\

\\
\#\#\# Response
\\
    \bottomrule
    \end{tabular}
    \caption{
    The template for zero-shot inference using the baseline LLMs on TAT-DQA datasets.}
    \label{tab:zero_tatdqa}
\end{table*}

\end{document}